\documentclass{article}

\PassOptionsToPackage{numbers,square,sort&compress}{natbib}
\usepackage{arxiv}

\usepackage{amsmath,amssymb,amsfonts}
\usepackage{graphicx}
\usepackage{booktabs}
\usepackage{makecell}
\usepackage{multirow}
\usepackage{array}
\usepackage{xcolor}
\usepackage{url}
\usepackage{algorithmic}
\usepackage[ruled,vlined,linesnumbered]{algorithm2e}
\usepackage{lineno}
\usepackage{hyperref}
\hypersetup{
	colorlinks=true,
	linkcolor=blue,
	citecolor=blue,
	urlcolor=blue
}

\newcommand{\blueeqref}[1]{\textcolor{blue}{\eqref{#1}}}

\begin{document}
	\title{Learning to Optimize UAV Path Planning for Data Sensing in Wireless Sensor Networks}
	
	\author{
        \textbf{Sijie~Ma}$^{1}$,
        \textbf{Zeyuan~Ma}$^{2,}$\thanks{Zeyuan Ma is the corresponding author (mzy@ieee.org).}~,
        \textbf{Weijia~Cao}$^{3}$\\
        \textbf{Yue-Jiao~Gong}$^{1}$,
        \textbf{Lingling~Ma}$^{3}$,
        \textbf{Zhiyang~Huang}$^{1}$,
        \textbf{Jun~Zhang}$^{4}$\\[-1pt]
        $^{1}$South China University of Technology \quad $^{2}$South China Normal University\\
        $^{3}$Aerospace Information Research Institute, CAS \quad $^{4}$Nankai University
	}
	
	\maketitle
	
	\begin{abstract}
		UAVs have emerged as highly flexible platforms for data sensing in Wireless Sensor Networks (WSNs). Path planning for UAVs in such tasks plays a key role to assure remote sensing effectiveness and friendly energy consumption. However, existing approaches show two key limitations: i) they are primarily hand-crafted with certain design biases that harm adaptation on unseen tasks. ii) they predominantly assume idealized spatial complexities of actual environments through simplified simulation, causing them to underperform during real-world deployment. In this paper, we propose a novel learning-assisted planning framework, termed Landscape-Aware Meta Differential Evolution (LAMDE), to tackle the mentioned limitations. The major contributions come from the following aspects. We first re-formulate such UAV path planning problem to embrace challenging constraints. To efficiently navigate this highly constrained space, we propose a bi-level learning to optimize approach, where the meta-level is a trainable algorithm configuration policy that meta-learns an adaptable planning strategy for low-level planning algorithm. To address the potential training data scarcity and distribution shift in real-world environments, we introduce a landscape-aware automatic augmentation scheme that enriches training data. At the low-level, a Differential Evolution algorithm is deployed for solving the path planning tasks. To enhance the solving flexibility, we further design a variable-length encoding strategy that dynamically prunes redundant hover points and optimizes continuous flight parameters concurrently within a unified search space. Based on all proposed designs, we meta-train LAMDE and compare it with representative baselines. Comprehensive experiments demonstrate that LAMDE achieves state-of-the-art performance on the tested complex UAV path planning tasks in WSN data collection scenarios. 
	\end{abstract}


	\section{Introduction}
	\label{sec:intro}
	Unmanned aerial vehicles~(UAVs) have been extensively deployed across diverse domains, including environmental monitoring~\cite{ahmad2025future,ya2017uav}, precision agriculture~\cite{zhang2025review}, disaster emergency response~\cite{daud2022applications}, and wireless sensor network (WSN) data collection~\cite{baek2019energy, zhan2017energy}. Among these applications, WSN data collection plays a vital role in industrial and smart infrastructure systems~\cite{poorvi2025reliable}. However, practical deployment remains challenging due to complex constraints such as energy consumption and obstacle environments~\cite{uav_tai_2, yang2018energy,jones2023path}.
	
	Evolutionary Computation (EC) methods are widely adopted for optimizing UAV data collection tasks~\cite{lin2020energy, shi2020uav}. However, current EC-based approaches still exhibit several inherent limitations when applied to complex real-world missions:
	1) \textbf{Oversimplified Modeling}: they often simplify the planning tasks by assuming obstacle-free environments~\cite{goudarzi2023uav, liu2024multi}; 2) \textbf{Inflexible Encoding}: they more or less  adopt a predefined, fixed number of hovering points~\cite{bai2025uav, sun20253d} for their solution encoding, which limits the optimization flexibility under realistic spatial constraints.
	3) \textbf{Human-crafted Design}: they predominantly adopt conventional human-crafted optimizers for the planning tasks, which rely on labor-intensive manual mechanisms and additional sensitive hyper-parameters, ultimately weakening robustness and generalization.
	
	Given the effectiveness and automated problem solving potential observed in recent Meta-Black-Box Optimization~(MetaBBO) works~\cite{metabbosurvey}, in this paper, we propose a holistic learning-assisted EC framework termed as Landscape-Aware Meta Differential Evolution (\textbf{LAMDE}), which introduces meta-learning paradigm~\cite{meta-learning} to learn effective and generalizable optimizer settings for automatic and end-to-end UAV path planning. At the meta-level, a reinforcement learning-based algorithm configuration policy is used to configure the low-level EC optimizer~\cite{de} in a dynamic manner. The policy is trained on a problem distribution to enhance the expexted planning performance. Following this paradigm, we propose several novel designs in LAMDE for addressing the limitations mentioned above:
	
	1) \textbf{Modeling Reformulation}: to better reflect practical challenges in real-world UAV data collection missions, we formulate a comprehensive UAV trajectory optimization model that explicitly incorporates complex physical constraints, including heterogeneous building geometries and customized no-fly zones. 
	
	2) \textbf{Variable-length Encoding}: to mitigate the downside of the dimension-fixed encoding in existing EC optimizers, we propose a variable-length encoding strategy that supports flexible dimensional representation for solution and enables automatic pruning of redundant trajectories, hence improving the flexibility and convergence in the low-level optimization.

	3) \textbf{Training Data Augmentation}: while MetaBBO successfully alleviates the manual design burdens of classical EC, they normally require a strict distributional alignment between the training problems and tested ones~\cite{abom}. This is impractical in our data collection tasks, since high-quality prior training data is profoundly scarce. To this end, we landscape-aware training data preparation scheme designed to circumvent the dependency on a predefined training distribution through landscape matching between the target UAV planning tasks and automatically generated synthetic problems. Such data augmentation scheme not only relieve the burden of real-world training data, but also ensures the distributional alignment between training and testing.
	
	Rigorous empirical validations show that LAMDE achieves state-of-the-art performance on the proposed UAV tasks compared to both traditional or learning-based EC optimizers. Further ablations and in-depth analysis underscore the effectiveness of each newly proposed design.

	\section{Related Work}
	\label{sec:related}
	\subsection{UAV Path Planning}
	In Wireless Sensor Network data collection, the modeling paradigm of UAV path planning largely determines the complexity and realism of the optimization task. Early studies typically formulate this problem as a discrete combinatorial optimization task, modeled as variants of the Traveling Salesman Problem (TSP), where the objective is to determine an efficient visiting order of sensor nodes~\cite{shivgan2020energy, joseph2020uav}. To enable smoother and more flexible trajectories, recent works have extended this formulation to continuous spaces, representing UAV paths using spatial coordinates or spline curves to balance flight distance, energy consumption, and data coverage~\cite{bai2025uav, sun20253d}. Despite the shift toward continuous spatial modeling, current paradigms remain limited by insufficiently realistic constraint modeling. Most studies either assume idealized obstacle-free environments~\cite{lin2020energy, shi2020uav} or rely on coarse grid-based approximations, which fail to accurately capture complex building geometries and no-fly zones. As a result, these formulations cannot faithfully reflect the spatial constraints encountered in real-world deployments, leading to a mismatch between modeled and practical scenarios. Therefore, developing a realistically constrained problem formulation that explicitly incorporates complex environments is essential for bridging this gap.
	
	Evolutionary Computation (EC) and swarm intelligence algorithms are widely adopted to solve the aforementioned UAV routing paradigms due to their global search capabilities and independence from derivative information~\cite{uav_tai_1}. For early discrete formulations, Genetic Algorithms (GAs) are often customized with specialized heuristics to handle routing tasks under strict time limits~\cite{joseph2020uav}. Ant Colony Optimization (ACO) is extensively utilized to construct routes with minimal energy consumption~\cite{lin2020energy} and is further enhanced via sequential decision strategies to optimize the Age of Information~\cite{gao2023aoi}. As research advances toward continuous spatial modeling, continuous optimizers such as Particle Swarm Optimization (PSO) and Differential Evolution (DE) have been widely adopted. PSO employs refined path representations to reduce flight costs~\cite{shi2020uav, liu2024aoi}, while DE variants such as Matrix Differential Evolution (MDE) have been proposed to better handle complex spatial coordinates~\cite{bai2025uav}. Although successful in idealized or coarsely approximated settings, conventional EC algorithms exhibit notable limitations in realistically constrained environments. A key issue is the requirement to predefine a fixed number of UAV hovering points before optimization, which imposes structural rigidity and prevents flexible trajectory adaptation in complex environments. In addition, their performance often relies heavily on handcrafted heuristics and extensive parameter tuning, making them difficult to transfer across different mission scenarios. As a result, their generalization ability in diverse and unpredictable landscapes is significantly restricted. Motivated by these challenges, this work introduces an automated MetaBBO framework for UAV path planning, featuring a tailored variable-length encoding strategy to enable the adaptive scaling of hovering points.
	
	\subsection{Meta-Black-Box Optimization}
	Back to 1987, meta-learning~\cite{schmidhuber1987evolutionary} was proposed to learn the credit assignment method itself through self-modifying code, e.g., self-referential Gödel machine~\cite{schmidhuber2007godel}. Following this principle, Meta-Black-Box Optimization (MetaBBO)~\cite{metabbosurvey, yang2025meta} adopts a bi-level learning framework to enhance the performance of underlying EC optimizers. In this paradigm, a meta-level agent is trained to guide or control the behavior of low-level optimization processes, enabling adaptive and data-driven search strategies. Considering the target algorithm design tasks, existing studies can generally be categorized into four progressively advanced paradigms: Algorithm Selection~\cite{rldas}, Algorithm Configuration~\cite{gleet}, Solution Manipulation~\cite{glhf} and Algorithm Generation~\cite{designx}. Given such a broad spectrum of algorithm design tasks, the meta learning in MetaBBO is instantiated with diverse training paradigms such as reinforcement learning~\cite{rl-bbo-survey}, (self-)supervised learning~\cite{abom}, neuroevolution~\cite{lange2023discovering} and LLM-based in-context learning~\cite{eoh}. Despite these advances, MetaBBO still faces a practical challenge in its learning: the distribution shift from the training problems to the testing ones. Most existing methods implicitly assume that the synthetic training problems align with the realistic problems to be tested. However, in real-world scenarios, the true task distribution is often difficult to characterize, or explicitly divergent from the training set. Our LAMDE is exactly motivated by this issue and address it by the novel landscape-aware data preparation module.
	
	\begin{figure}[t]
		\centering
		\includegraphics[width=0.7\linewidth]{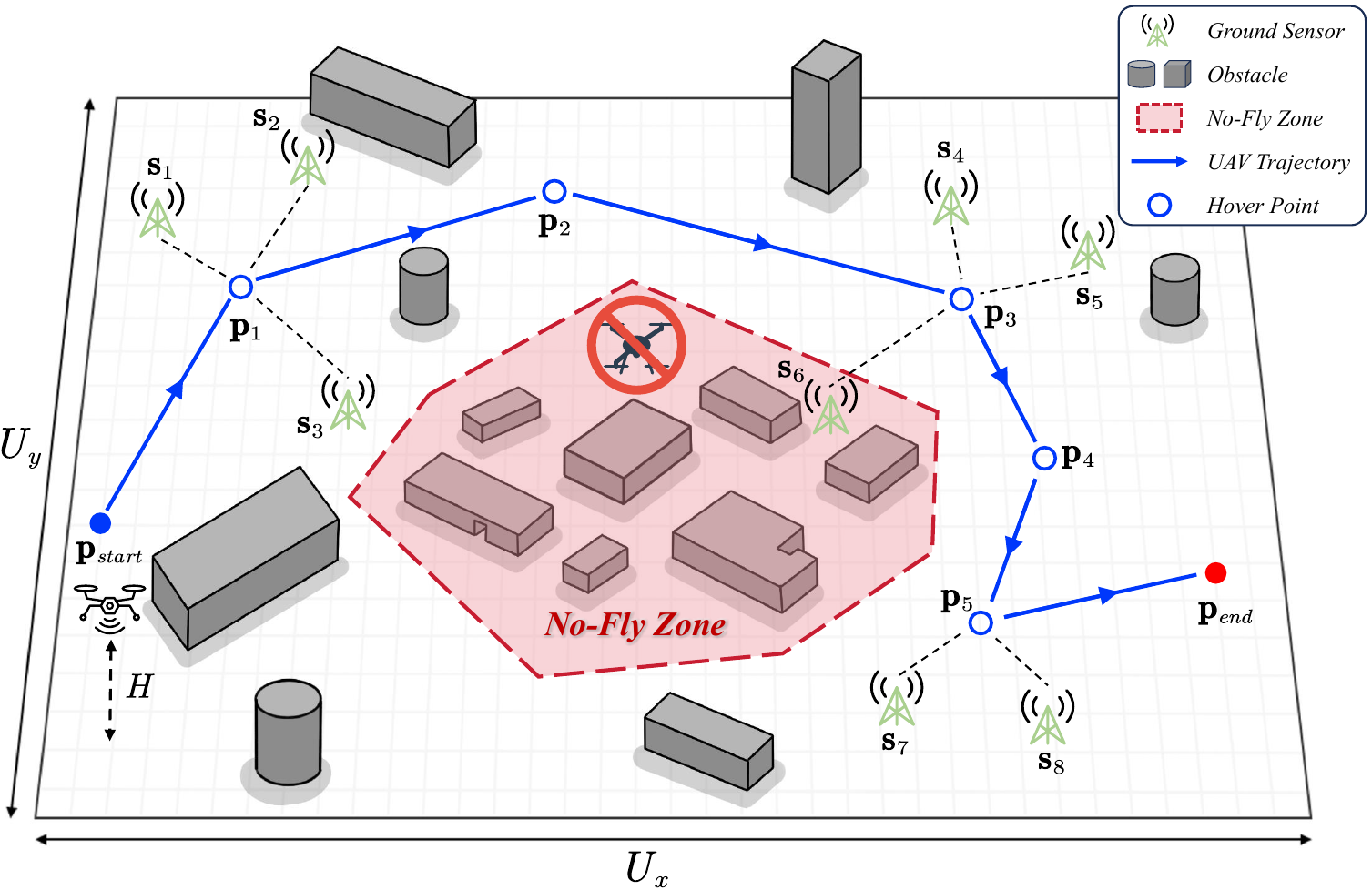}
		\caption{Overview of UAV path planning for ground sensor data collection under environmental constraints, including obstacles and no-fly zones.}
		\label{fig:problem}
	\end{figure}
	
	\section{Problem Definition and Modeling}
	\label{sec:problem_formulation}
	\subsection{UAV Path Planning in WSN Data Collection}
	\label{subsec:basic_model}
	
	In WSN data collection tasks, the aim is using UAV to fly through the sensor areas and collect data as much as possible. A recent work~\cite{bai2025uav} provides a simplified and clear path planning problem definition for such tasks. Based on that, we re-formulate a more practical UAV-enabled data collection model by explicitly incorporating complex environmental constraints. We consider a rotary-wing UAV collecting data from a group of $J$ ground sensors $\mathbf{S}=\{\mathbf{s}_1, \dots, \mathbf{s}_J\}$ distributed over a target area $\mathcal{A} = [0, U_x] \times [0, U_y]$. To reduce energy consumption, the UAV operates at a constant cruising altitude $H$, which simplifies trajectory planning to a 2D spatial problem. As illustrated in Fig.~\ref{fig:problem}, the UAV travels from a predefined start point $\mathbf{p}_{start}$ to an end destination $\mathbf{p}_{end}$, while collecting data from sensors and avoiding obstacles and no-fly zones. The trajectory is denoted as $\{\mathbf{p}_0, \mathbf{p}_1, \dots, \mathbf{p}_I, \mathbf{p}_{I+1}\}$. The intermediate set $\mathbf{P} = \{\mathbf{p}_1, \dots, \mathbf{p}_I\}$ defines the planned UAV path. 
	
	The UAV operates in two alternating phases: flying between hover points and hovering to collect data from the sensors. During the flying phase between consecutive points $\mathbf{p}_i$ and $\mathbf{p}_{i+1}$, the UAV follows a trapezoidal velocity profile with maximum speed $v_{max}$ and constant acceleration $a$. The flight time $T_{fly,i}$ for distance $d_i = \|\mathbf{p}_{i+1} - \mathbf{p}_i\|_2$ is given by:
	\begin{equation}
		T_{fly,i} = 
		\begin{cases} 
			\frac{d_i}{v_{max}} + \frac{v_{max}}{a}, & \text{if } d_i > \frac{v_{max}^2}{a} \\
			2\sqrt{\frac{d_i}{a}}, & \text{if } d_i \le \frac{v_{max}^2}{a}
		\end{cases}
		\label{eq:flight_time}
	\end{equation}
	
	Data collection occurs strictly during the hovering phase. The data transmission rate $R_{i,j}$ between the UAV hovering at $\mathbf{p}_i$ and the $j$-th sensor at $\mathbf{s}_j$ is modeled as $R_{i,j} = \max \left\{ R_c - \xi \cdot \sqrt{\|\mathbf{p}_i - \mathbf{s}_j\|_2^2 + H^2}, 0 \right\}$, where $R_c$ is the maximum theoretical data rate and $\xi$ is the signal decay factor. Here we note that the $\max$ operation indicates that if a hover point is too distant from a sensor, no data could be collected. Consequently, for a sensor $j$, the overall data collected during the flying trajectory $\mathbf{P}$ is aggregated as $Q_j = \sum_{i=1}^{I} R_{i,j} \cdot T_{hov,i}$, where $\mathbf{T}_{hov} = \{T_{hov,1}, \dots, T_{hov,I}\}$ are the hover duration times on the hover points.
	
	In such path planning setting, the decision variables are the union of flying trajectory and its corresponding hover duration times: $\mathbf{x} = \{\mathbf{P}, \mathbf{T}_{hov}\}$. We will discuss how $\mathbf{x}$ participates in the final optimization model in Sec.~\ref{subsec:overall_formulation}.
	
	\subsection{Environmental Modeling with Obstacles}
	\label{subsec:env_modeling}
	Based on the basic path planning setting above, we further extend it with physical obstructions and regulatory boundaries commonly found in urban or suburban areas, so as to bridge the gap between theoretical modeling and real-world deployments. 
	
	Specifically, we enhance the simulation's realism by introducing a comprehensive and heterogeneous set of static constraints $\mathcal{O}$. This set is formulated to include multiple circular structures ($\mathcal{O}_{cir}$), rotated rectangular buildings ($\mathcal{O}_{rec}$), and irregular No-Fly Zones ($\mathcal{O}_{nfz}$). The specific geometric definitions of these obstacles, which introduce significant non-convexity into the problem space, are detailed as follows:
	
	\noindent\textbf{Circular Obstacles.} Symmetric structures such as water towers or signal stations are modeled as a set of circular obstacles $\mathcal{O}_{cir}$. The $k$-th obstacle in this set is defined as a circular area of which the center is $\mathbf{c}_k = [x_k, y_k]^\top$ and radius is $r_k$. To reflect the reality, $r_k$ is with a range of 40-100m.
	
	\noindent\textbf{Rectangular Obstacles.} Residential and commercial buildings are modeled as a set of rectangular obstacles $\mathcal{O}_{rec}$. To reflect realistic urban layouts rather than simplified axis-aligned grids, our model considers arbitrary orientations. Under such consideration, each rectangular obstacle in $\mathcal{O}_{rec}$ is defined by its geometric center $\mathbf{c} = [x_c, y_c]^\top$, length $L$, width $W$, and rotation angle $\theta \in [0, \pi)$. $L$ and $W$ are with the range of 80-180m.  
	
	\noindent\textbf{No-Fly Zones (NFZs).} Large-scale restricted areas, such as industrial parks, are introduced as No-Fly Zones set $\mathcal{O}_{nfz}$. Each zone in this is geometrically modeled as a convex polygon defined by an ordered sequence of $K$ vertices $\{\mathbf{v}_{1}, \dots, \mathbf{v}_{K}\}$ arranged in counter-clockwise order. $K$ is with the range of 3-7, the diameter of the convex polygon is between 3000-4000m. 
	
	We note that in the simulation of the experiments, we randomly generate these three types of obstacles to construct each obstacle set, ensuring they are distributed following reasonable landscapes in reality. Due to the space limitation, we leave detailed protocols in our project.  
	
	\subsection{Modeling of Collision Risk}
	\label{subsec:safety constraints}
	With the challenging flying environment with multiple obstacles, we need a metric to measure the collision risk of a flying trajectory $\mathbf{P}$ under a given map:  $L_{vio}(\mathbf{P})$. Following the first principle, we use a straightforward way to compute $L_{vio}(\mathbf{P})$:
	\begin{equation}
		\label{eq:L_vio}
		L_{vio}(\mathbf{P}) = 
		L_{rec}(\mathbf{P}) 
		+ L_{cir}(\mathbf{P}) 
		+ L_{nfz}(\mathbf{P})
		,
	\end{equation}
	where $L_{rec}(\mathbf{P})$, $L_{cir}(\mathbf{P})$ and $L_{nfz}(\mathbf{P})$ are the total length of the overlapped part between the flying trajectory and each obstacle set respectively. $L_{vio}(\mathbf{P})$ serves as an important constraint in our final optimization model.
	
	\subsection{Optimization Model}
	\label{subsec:overall_formulation}
	With all necessary components established, we can now provide the final optimization model used in this paper. The objective is to minimize a weighted sum of total flight time and hovering time, balancing mission efficiency and energy consumption. Through integrating the UAV dynamics, data requirements, and geometric safety constraints, the overall path planning problem can be formulated as:
	\begin{equation}
		\label{eq:problem}
		\begin{aligned}
			\min_{\mathbf{x}} \quad & f(\mathbf{x}) = \alpha \cdot \sum_{i=0}^{I} T_{fly,i} + (1-\alpha) \cdot \sum_{i=1}^{I} T_{hov,i}, \\
			\text{s.t.} \quad & \left\{
			\begin{aligned}
				& \mathbf{p}_i \in \mathcal{A}, && \forall i \in \{1, \dots, I\} && \text{(C1)} \\
				& 0 \le T_{hov,i} \le T_{max}, && \forall i \in \{1, \dots, I\} && \text{(C2)} \\
				& \sum_{j=1}^{J} R_{i,j} \le R_{\lambda}, && \forall i \in \{1, \dots, I\} && \text{(C3)} \\
				& Q_j \ge Q_{\lambda}, && \forall j \in \{1, \dots, J\} && \text{(C4)} \\
				& L_{vio}(\mathbf{P}) = 0, && && \text{(C5)}
			\end{aligned}
			\right.
		\end{aligned}
	\end{equation}
	
	Here, $\mathbf{x} = \{\mathbf{P}, \mathbf{T}_{hov}\}$ represents the optimization variables, including the planned path $\mathbf{P}$ and the corresponding hovering durations $\mathbf{T}_{hov}$. In the objective function, $\alpha \in [0,1]$ is the weight factor that controls the trade-off between the flight time calculated in Eq.~\blueeqref{eq:flight_time} and the hovering time. Constraints (C1) and (C2) restrict the hover points to the target area and impose operational duration limits. (C3) guarantees the instantaneous data transfer rate respects the communication bandwidth $R_{\lambda}$, while (C4) ensures the total collected data $Q_j$ meets the mission requirement $Q_{\lambda}$. Finally, (C5) guarantees a strictly collision-free trajectory by completely nullifying the penalty term $L_{vio}(\mathbf{P})$ defined in Eq.~\blueeqref{eq:L_vio}. Designing an effective algorithm by hand for solving such complex constrained optimization problems is challenging. To this end, we propose LAMDE to automate the algorithm design by meta learning, which will be detailed in the next section.
	
	\section{Methodology}
	\label{sec:methodology}
	
	\subsection{Low-Level Optimizer}
	\label{sec:lower_level}
	The low-level optimizer is a Differential Evolution (DE) algorithm with tailored designs to support flexible search behavior, smooth constraint handling, and effective control by the meta-level RL policy.
	
	\noindent\textbf{Variable-Length Encoding Strategy.}\label{subsubsec:encoding} Existing works that address UAV path planning tasks lean to simplify the solution encoding, that is, fix the number of hover points as a prior by human experts~\cite{bai2025uav,sun20253d}. Such setting may benefit the algorithm complexity aspect, however, may also result in ineffective task solving. To this end, we consider a relatively large maximum number of allowed hover points $I_{\max}$\footnote{We use a simple heuristic: $I_{\max}=10$ for a 5 km $\times$ 5 km map and $I_{\max}=20$ for a 10 km $\times$ 10 km map. Future users can also modify this value by experiences.}. Furthermore, we add an additional dimension into the decision variables $\mathbf{x} = \{\mathbf{P}, \mathbf{T}_{hov}\}$ in our optimization model, which is the activation value $\epsilon$ for each hover point. The resulting candidate solution encoding is organized as:
	\begin{equation}
		\mathbf{x}=\left[\mathbf{u}_{1},\mathbf{u}_{2},\dots,\mathbf{u}_{I_{\max}}\right],
	\end{equation}
	where each $\mathbf{u}_{i}=[x_{i},y_{i},t_{i},\epsilon_{i}]$ represents a candidate hover point. These variables are bounded by $x_{i} \in [0, U_x]$, $y_{i} \in [0, U_y]$, $t_{i} \in [0, T_{\max}]$, and $\epsilon_{i} \in [0, 1]$. 
	
	We decode $\mathbf{x}$ into a practical flying plan through the following condition-based selection: 
	\begin{equation}
		\left\{(x_{i},y_{i},t_{i})\,\middle|\,\epsilon_{i}\ge\tau,\;i=1,\dots,I_{\max}\right\},
	\end{equation}
	where $\tau$ is a predefined activation threshold. We only keep the hover points whose activation value $\epsilon_i$ larger than the threshold as the actual hover points in the final flying plan. We abuse $I$ as the number of these points in following sections. Note that $\tau$ is an adjustable value (0.5 by default): a larger $\tau$ results in sparse flying plan and vice versa.
	
	\noindent\textbf{Constraint Handling Design.}\label{subsubsec:constraint_handling} In our optimization model, a solution needs to satisfy intricate constraints. To quantify constraint violations during optimization, we introduce explicit penalty metrics for constraint measuring. Recall the constraints we defined in Eq.~\ref{eq:problem}, the basic spatial and temporal constraints (C1) and (C2) are enforced to be satisfied since our DE optimizer will do value clamping operation on the decision variables. Hence, we mainly consider the violation status of the rest three constraints. Specifically, violations for communication bandwidth (C3) and data collection (C4) are defined as
	$\mathcal{V}_{C3,i} = \max \{ 0, \sum_{j=1}^{J} R_{i,j} - R_{\lambda} \}$ and 
	$\mathcal{V}_{C4,j} = \max \{ 0, Q_{\lambda} - Q_j \}$, respectively. For the constraint (C5), we define the violation based on Eq.~\blueeqref{eq:L_vio} as:
	\begin{equation}
		\mathcal{V}_{C5} = 
		\begin{cases}
			\mu + \nu \cdot L_{vio}(\mathbf{P}), & \text{if } L_{vio}(\mathbf{P}) > 0 \\
			0, & \text{otherwise}
		\end{cases}
	\end{equation}
	where $\mathbf{P}$ is decoded from $\mathbf{x}$, $\mu$ and $\nu$ are normalization factors that play a key role in stabilizing the relative quantity. The violation of (C5) is much larger than those of (C3) and (C4), since (C5) measures the overlap length between the flying trajectory and obstacles. This helps re-balance the attention of the DE optimizer on these constraints during the low-level optimization. We leave the values of $\mu$ and $\nu$ in the experiments. 
	The total constraint violation $\phi(\mathbf{x})$ is then aggregated:
	\begin{equation}
		\phi(\mathbf{x}) = \sum_{i=1}^{I} \mathcal{V}_{C3,i} + \sum_{j=1}^{J} \mathcal{V}_{C4,j} +  \mathcal{V}_{C5}.
	\end{equation}
	This unified scalar metric $\phi(\mathbf{x})$ serves as two roles in LAMDE: 1) it controls the selection pressure in the low-level DE optimizer; 2) it contributes to the optimization state features used by the meta-level algorithm configuration policy.
	
	\noindent\textbf{Differential Evolution Operations and Selection.}\label{subsubsec:de_ops} The low-level optimization process is modeled as a Markov Decision Process~(MDP). A DE optimizer is used to optimize the UAV path planning tasks. For a given task, at each generation $g$, the DE's mutation and crossover operations are controlled by the meta-level algorithm configuration policy. This is achieved by several steps. We first extract an optimization state from current generation of solution population $\mathbf{X}^{(g)} = \{\mathbf{x}^{(g)}_i,\dots,\mathbf{x}^{(g)}_N\}$ where $N$ is the population size. Then the meta-level policy outputs an action by conditioning this optimization state. The action is then decoded as the concrete mutation/crossover operators and corresponding hyper-parameter values, which are used for current optimization generation. Then the population is updated toward offspring $\tilde{\mathbf{X}}^{(g)} = \{\tilde{\mathbf{x}}_1^{(g)}, \dots, \tilde{\mathbf{x}}_N^{(g)}\}$. 
	
	After the offspring are reproduced, our DE optimizer selects the population for $g+1$-th generation by Deb's feasibility rule:
	\begin{equation}
		\mathbf{x}_i^{(g+1)} =
		\begin{cases}
			\tilde{\mathbf{x}}_i^{(g)}, & \text{if } (\tilde{\phi}_i - \phi_i < 0 )  \\
			\tilde{\mathbf{x}}_i^{(g)}, & \text{if } (\tilde{\phi}_i - \phi_i = 0 \text{ and }  \tilde{f}_i < f_i) \\
			\mathbf{x}_i^{(g)}, & \text{otherwise}
		\end{cases},
	\end{equation}
	where $f_i$ and $\tilde{f}_i$ denote the objective values of $\mathbf{x}_i^{(g)}$ and $\tilde{\mathbf{x}}_i^{(g)}$ respectively, $\phi_i$ and $\tilde{\phi}_i$ denote the constraint violations of $\mathbf{x}_i^{(g)}$ and $\tilde{\mathbf{x}}_i^{(g)}$ respectively. By such selection procedure, the DE optimizer prioritizes constraint satisfaction over objective minimization, deterministically steering the population toward the feasible manifold.
	
	\subsection{Meta-Level Policy}
	\label{sec:meta_level}
	As mentioned above, we model the meta-level dynamic algorithm configuration of the low-level DE optimizer as MDP, where a meta-level policy~(a learnable neural network) adaptively dictates proper algorithmic configurations based on the optimization state of the low-level optimization progress. In this section, we clarify the details of the MDP design components, the neural network architecture and data flow of the meta-level policy, and the training workflow of the meta-level learning.
	
	\noindent\textbf{State.}\label{subsec:state} At generation $g$, the observed state comprises the population $\mathbf{X}^{(g)}$, objectives $f(\mathbf{X}^{(g)})$, violations $\phi(\mathbf{X}^{(g)})$, and search progress $g/G$. To ensure numerical stability, $\bar{\mathbf{X}}^{(g)} \in [0,1]$ is used as the normalized form of $\mathbf{X}^{(g)}$. Simultaneously, to resolve the severe magnitude disparities of $f$ and $\phi$ across heterogeneous tasks, we introduce a mantissa-exponent decomposition. Each scalar evaluation $y \in \{f, \phi\}$ is formulated as $y = \omega \times 10^b$, where $\omega \in [-1,1]$ is the continuous mantissa and $b \in \mathbb{Z}$ is the discrete exponent. By scaling the exponent as $\epsilon = b/\eta$ with constant $\eta$, the raw evaluations for the $i$-th individual are stabilized into tuples $(\omega_{f,i}^{(g)}, \epsilon_{f,i}^{(g)})$ and $(\omega_{\phi,i}^{(g)}, \epsilon_{\phi,i}^{(g)})$. The optimization state at the $g$-th generation $s_g$ is defined as:
	\begin{equation}
		\label{eq:state}
		s_g = \left\{ \bar{\mathbf{X}}^{(g)}, \{(\omega_{f,i}^{(g)}, \epsilon_{f,i}^{(g)})\}_{i=1}^N, \{(\omega_{\phi,i}^{(g)}, \epsilon_{\phi,i}^{(g)})\}_{i=1}^N, \frac{g}{G} \right\}.
	\end{equation}
	
	\noindent\textbf{Action.}\label{subsec:action} At each generation $g$, the meta-level policy $\pi_\theta$ outputs a joint action $a_g$ that specifies the evolutionary configuration for each individual in the low-level DE's population (see supplementary materials for the detailed architecture of $\pi_\theta$). To enable fine-grained control, the action space is formulated as a hybrid of discrete operator selections and continuous hyper-parameter assignments:
	\begin{equation}
		\label{eq:action}
		a_g = \pi_\theta(s_g) = \left\{ \left( a_{i,g}^{os,m}, a_{i,g}^{os,c}, F_{i,g}, Cr_{i,g} \right) \right\}_{i=1}^N,
	\end{equation}
	where the number of actions for the whole population is $N\times 4$. For the $i$-th individual, the 4 actions can be categorized into two classes:
	\begin{itemize}
		\item \textbf{Discrete Operator Selection:} The categorical variable $a_{i,g}^{os,m} \in \{1,\dots,7\}$ selects the mutation strategy from seven classical variants (\textit{rand/1}, \textit{best/1}, \textit{rand/2}, \textit{best/2}, \textit{current-to-rand/1}, \textit{current-to-best/1}, \textit{rand-to-best/1}). 
		Similarly, $a_{i,g}^{os,c} \in \{1,2\}$ determines the crossover scheme, choosing between \textit{binomial} and \textit{exponential}.
		
		\item \textbf{Continuous Parameter Assignment:} The continuous variables $F_{i,g} \in [0,1]$ and $Cr_{i,g} \in [0,1]$ specify the scaling factor and crossover rate for the selected operators.
	\end{itemize}
	
	Given the action $a_g$, the low-level DE optimizer updates the population by executing the corresponding mutation and crossover operators to reproduce the offspring, and then  selection operation to determine the next generation population, as detailed in Section~\ref{subsubsec:de_ops}.
	
	\noindent\textbf{Reward.}\label{subsec:reward} Upon completing the population update, the low-level optimization environment returns a scalar reward $r_g$ to the meta-level to guide the policy's learning. Let $\langle\phi_{best}^{(g)}, f_{best}^{(g)}\rangle$ denote the constraint violation and objective value of the same best individual at generation $g$. The reward $r_g$ is defined as:
	\begin{equation}
		\label{eq:reward}
		r_g =
		\begin{cases}
			\frac{\phi_{best}^{(g-1)} - \phi_{best}^{(g)}}{\phi_{best}^{(g-1)}}, & \text{if } \phi_{best}^{(g)} > \delta \\
			1 + \exp\left(-\frac{g}{G}\right), & \text{if } \phi_{best}^{(g-1)} > \delta \ge \phi_{best}^{(g)} \\
			\frac{f_{best}^{(g-1)} - f_{best}^{(g)}}{f_{best}^{(g-1)}}, & \text{if } \phi_{best}^{(g-1)} \le \delta \text{ and } \phi_{best}^{(g)} \le \delta
		\end{cases}
	\end{equation}
	where $\delta$ is a predefined feasibility tolerance. This reward design encourages constraint violation reduction in the early stage, provides an additional bonus when entering the feasible region, and focuses on objective improvement thereafter.

	\subsection{Training}
	\label{sec:training}
	This section presents the training procedure of LAMDE. As an initial exploration on deploying MetaBBO for optimization tasks in the wild, LAMDE faces a key challenge, which is that the absence of a well-defined training distribution~(since many realistic scenarios lack sufficient problem instances) makes it difficult to obtain sufficient training instances for LAMDE to learn. While existing MetaBBO works claim that a subtle training on diverse synthetic benchmarks~(e.g., CoCo-BBOB benchmark~\cite{bbob} and IEEE CEC BBO benchmark~\cite{liang2014problem}) is sufficient to ensure generalization, it is not the case when we do preliminary validation on the UAV path planning tasks in this paper. We believe this is due to the complex optimization structures in such realistic scenarios, which results in significant distribution shift. In the following sections, we will first introduce how we address this training issue and then provide details on the overall training workflow.
	
	\noindent\textbf{Landscape Aware Module.}\label{sec:landscape-aware} To address the potential training distribution shift issue, we introduce a simple yet effective scheme to automatically construct landscape-similar synthetic training problems for UAV path planning problems in this paper. We term this scheme as \textit{Landscape-Aware Module}. Given a group of target realistic optimization tasks we want to solve (i.e., the test set): $\mathcal{D}_{\mathrm{test}}=\{\mathcal{T}_{1},\dots,\mathcal{T}_{M}\}$, for the $m$-th realistic task, it is defined by its objective function and constraints: $\mathcal{T}_m = \langle f_m, \phi_m \rangle$. For $\mathcal{D}_{\mathrm{test}}$, our landscape-aware module regards it as a target problem distribution and is capable of finding moderate landscape-similar synthetic problems from a large synthetic problem database. Then the found problems constitute a training problem set with sufficient training instances and more importantly, distributional similarity with $\mathcal{D}_{\mathrm{test}}$.  
	
	To achieve this, we need to construct a synthetic problem database $\mathcal{D}_{syn}$ first. Fortunately, the representative CoCo-BBOB benchmark~\cite{bbob} provides us a group of 24 basis synthetic functions with diverse optimization properties. Based on it, we use the problem augmentation interface in a recent learning-based optimization benchmark platform, MetaBox~\cite{metabox,metaboxv2}, to augment these 24 functions as a database $\mathcal{D}_{syn}$. This interface supports us to set different problem dimensions and random shift/rotation transformation for the 24 basis functions, resulting in 14400 diverse BBO instances\footnote{For each basis function, we select problem dimensions from $\{20,40,80,120,160,200\}$, and generate 100 instances using random shift and rotation, resulting in $24\times 6\times 100=14400$ instances.}. With the diversity-enhanced synthetic problem database, we next introduce how we query landscape-similar instances for the target realistic optimization tasks in $\mathcal{D}_{\mathrm{test}}$. 
	
	Generally speaking, we leverage landscape feature similarity matching to identify similar synthetic instances for target realistic tasks. Specifically, we probe the underlying landscape space of the tasks in $\mathcal{D}_{\mathrm{test}}$ by first using a Latin Hypercube Sampling (LHS) to sample $N_{\mathrm{probe}}$ points $\mathbf{X}_m$ from the solution space. Then we again utilize the NeurELA feature extractor $\Psi(\cdot)$ to independently extract two groups of landscape features from the objective landscape and constraint violation space: $\mathbf{z}_{m}^{f}=\Psi(\mathbf{X}_m, f_m(\mathbf{X}_m))$ and $\mathbf{z}_{m}^{\phi}=\Psi(\mathbf{X}_m,\phi_m(\mathbf{X}_m))$. For each synthetic instance $s$ in the synthetic problem database $\mathcal{D}_{syn}$, we follow the same feature extraction procedure and denote the attained feature as $\mathbf{z}_s$. Given the feature vectors computed offline, we establish a basis for measuring similarity between problems in landscape feature space.
	
	We next propose a two-stage query scheme to fill up $\mathcal{D}_{\mathrm{train}}$. This is achieved by first locating a subset $\mathcal{S}_{f}$ of $\mathcal{D}_{syn}$, which comprises $K$~($K > M$) different instances that show nearest distributional distance to the tasks in $\mathcal{D}_{\mathrm{test}}$:
	\begin{equation}
		\label{eq:select_obj}
		\mathcal{S}_{f}=
		\underset{\mathcal{S}\subset\mathcal{D}_{syn},\,|\mathcal{S}|=K}{\arg\min}
		\sum_{s\in\mathcal{S}}
		\left(\frac{1}{M}\sum_{m=1}^{M}
		\left\| \mathbf{z}_m^f - \mathbf{z}_s\right\|_2\right).
	\end{equation}
	In the second stage, for each located objective function $s_k \in \mathcal{S}$, we focus on locating a best-matching synthetic instance in $\mathcal{D}_{syn}$ that holds the same problem dimension as $s$ and is the nearest to the corresponding constraint feature $\mathbf{z}_{m}^{\phi}$. We denote this instance as $s^{\prime}_{k}$. Finally, the augmented training problem set $\mathcal{D}_{\mathrm{train}}$ is constructed by:
	\begin{equation}
		\mathcal{D}_{\mathrm{train}} = \{<s_k, s^{\prime}_{k}>\}_{k=1}^K,
	\end{equation}
	which comprises $K$ training problem instances. For each training problem $<s_k, s^{\prime}_{k}>$ in $\mathcal{D}_{\mathrm{train}}$, the evaluation is kept in a simple way: given a solution $\mathbf{x}$, its objective value is $s_k(\mathbf{x})$, and its constraint violation measure $\phi(\mathbf{x}) = s^{\prime}_{k}(\mathbf{x})$. LAMDE is trained on our constructed proxy set $\mathcal{D}_{\mathrm{train}}$ to enhance the optimization performance on the target task set $\mathcal{D}_{\mathrm{test}}$. By doing so, even if we have limited realistic task instances in the UAV path planning scenario, we could augment sufficient training data by our proposed scheme. This is a key contribution to existing MetaBBO works to relieve the distribution shift and data limitation.
	
	\begin{algorithm}[t]\small
		\caption{Training Procedure of LAMDE}
		\label{alg:framework}
		
		\SetKwInOut{Input}{Input}
		\SetKwInOut{Output}{Output}
		\SetKwComment{Comment}{// }{}
		
		\Input{Policy $\pi_{\theta}$, critic $V_\psi$, $\mathcal{D}_{\mathrm{test}}$, $\mathcal{D}_{\mathrm{syn}}$, population size $N$, generations $G$, max epoch $E$, PPO interval $n$, PPO epochs $\kappa$.}
		\Comment{Phase 1: Construct Training Set}
		$\mathcal{D}_{\mathrm{train}} \gets \mathrm{LAM}(\mathcal{D}_{\mathrm{test}}, \mathcal{D}_{\mathrm{syn}})$ (Sec.~\ref{sec:landscape-aware})\;
		\vspace{1mm}
		\Comment{Phase 2: Meta-Training via n-step PPO}
		\For{$epoch = 1, \dots, E$}{
			\For{$\mathcal{T} \in \mathcal{D}_{\mathrm{train}}$}{
				Initialize population $\mathbf{X}^{(1)}$, rollout buffer $\mathcal{B} \gets \emptyset$\;
				Observe initial state $s_1$ using Eq.~\blueeqref{eq:state}\;
				\For{$g = 1, \dots, G-1$}{
					Sample actions $a_g$ via Eq.~\blueeqref{eq:action}\;
					
					Evolve $\mathbf{X}^{(g)}$ to $\mathbf{X}^{(g+1)}$ by DE with $a_g$\;
					
					Observe next state $s_{g+1}$\;
					
					Evaluate reward $r_g$ via Eq.~\blueeqref{eq:reward}\;
					
					Update $\mathcal{B} \gets \mathcal{B} \cup \{(s_g, a_g, r_g, s_{g+1})\}$\;
					\If{$g \bmod n = 0$}{
						\For{$k = 1, \dots, \kappa$}{
							Update $\pi_\theta, V_\psi$ via PPO on $\mathcal{B}$\;
						}
						Clear $\mathcal{B} \gets \emptyset$\;
					}
				}
			}
		}
		
		\Output{Optimal meta-level policy $\pi_{\theta^*}$.}
	\end{algorithm}
	
	\noindent\textbf{Training Workflow.}\label{subsubsec:training_workflow}
	We summarize in Algorithm~\ref{alg:framework} the training procedure of the LAMDE. We adopt Proximal Policy Optimization (PPO) algorithm~\cite{ppo} to meta-train the meta-level policy. In lines 1, we first construct the training problem set as elaborated in Sec.~\ref{sec:landscape-aware}. Then in lines 2-15, PPO is used for the training process. For each epoch and each task, the low-level DE optimizer initializes a solution population and iteratively optimizes $\mathcal{T}$ for $G$ generations. At each generation $g$, the meta-level policy observes the state $s_g$ and outputs the action $a_g$. Guided by this action, the DE optimizer evolves the population forward, yielding the subsequent state $s_{g+1}$ and the improvement reward $r_g$. In lines 12-15, the actor and critic of the meta-level policy are updated by PPO's n-step update fashion. After the training, LAMDE could be used for solving unseen tasks in a zero-shot manner.

	\begin{table}[!t]
		\centering
		\caption{Results (mean $\pm$ std) of different algorithms. Symbols ``$+$'', ``$\approx$'', and ``$-$'' denote worse, similar, and better performance than LAMDE under the Wilcoxon rank-sum test ($95\%$). Best results are \textbf{bolded} and second-best are \underline{underlined}.}
		\label{tab:exp2}
		\scriptsize
		\setlength{\tabcolsep}{2pt}
		\resizebox{\textwidth}{!}{%
			\begin{tabular*}{\textwidth}{@{\extracolsep{\fill}} c *{9}{c} @{}}
				\toprule
				& \multicolumn{4}{c}{\textbf{BBO}} 
				& \multicolumn{4}{c}{\textbf{MetaBBO}} 
				& \textbf{Ours} \\
				\cmidrule(lr){2-5} \cmidrule(lr){6-9} \cmidrule(lr){10-10}
				Task & CMOCSO & CMA-ES & L-SRTDE & MDE-CGO & LDE & GLEET & RLDEAFL & ABOM & LAMDE \\
				\midrule
				
				$\mathcal{T}_{1}$ &
				\makecell{2.516e+3 \\ $\pm$3.847e+2} &
				\makecell{1.599e+3 \\ $\pm$7.334e+2} &
				\makecell{2.045e+3 \\ $\pm$4.904e+2} &
				\underline{\makecell{1.545e+3 \\ $\pm$8.951e+2}} &
				\makecell{2.365e+3 \\ $\pm$8.924e+1} &
				\makecell{2.439e+3 \\ $\pm$3.967e+2} &
				\makecell{1.641e+3 \\ $\pm$7.572e+2} &
				\makecell{1.603e+3 \\ $\pm$7.708e+2} &
				\textbf{\makecell{1.356e+3 \\ $\pm$7.791e+2}} \\
				
				$\mathcal{T}_{2}$ &
				\makecell{1.899e+3 \\ $\pm$1.068e+3} &
				\makecell{1.551e+3 \\ $\pm$8.877e+2} &
				\makecell{5.258e+2 \\ $\pm$3.370e+2} &
				\makecell{1.008e+3 \\ $\pm$7.196e+2} &
				\makecell{1.763e+3 \\ $\pm$9.227e+2} &
				\makecell{2.324e+3 \\ $\pm$8.434e+2} &
				\makecell{8.606e+2 \\ $\pm$8.364e+2} &
				\underline{\makecell{3.579e+2 \\ $\pm$3.054e+1}} &
				\textbf{\makecell{1.823e+2 \\ $\pm$7.575e+0}} \\
				
				$\mathcal{T}_{3}$ &
				\makecell{6.020e+3 \\ $\pm$1.507e+3} &
				\makecell{2.089e+3 \\ $\pm$1.382e+3} &
				\makecell{2.775e+3 \\ $\pm$9.554e+2} &
				\makecell{2.304e+3 \\ $\pm$1.287e+3} &
				\makecell{3.974e+3 \\ $\pm$8.768e+2} &
				\makecell{6.559e+3 \\ $\pm$9.026e+2} &
				\makecell{1.891e+3 \\ $\pm$1.011e+3} &
				\underline{\makecell{1.790e+3 \\ $\pm$7.003e+2}} &
				\textbf{\makecell{5.513e+2 \\ $\pm$6.686e+1}} \\
				
				$\mathcal{T}_{4}$ &
				\makecell{4.375e+3 \\ $\pm$4.065e+2} &
				\textbf{\makecell{1.973e+3 \\ $\pm$8.551e+2}} &
				\makecell{4.056e+3 \\ $\pm$1.971e+2} &
				\makecell{2.949e+3 \\ $\pm$8.856e+2} &
				\makecell{4.348e+3 \\ $\pm$2.567e+2} &
				\makecell{4.395e+3 \\ $\pm$2.851e+2} &
				\makecell{2.944e+3 \\ $\pm$4.428e+2} &
				\makecell{3.523e+3 \\ $\pm$2.295e+2} &
				\underline{\makecell{2.394e+3 \\ $\pm$6.119e+2}} \\

				$\mathcal{T}_{5}$ &
				\makecell{4.695e+2 \\ $\pm$9.505e+1} &
				\underline{\makecell{2.469e+2 \\ $\pm$1.716e+1}} &
				\makecell{3.563e+2 \\ $\pm$2.149e+1} &
				\makecell{4.624e+2 \\ $\pm$6.198e+1} &
				\makecell{4.829e+2 \\ $\pm$3.763e+1} &
				\makecell{4.419e+2 \\ $\pm$1.102e+2} &
				\makecell{3.091e+2 \\ $\pm$4.384e+1} &
				\makecell{3.252e+2 \\ $\pm$2.566e+1} &
				\textbf{\makecell{1.926e+2 \\ $\pm$5.952e+0}} \\
				$\mathcal{T}_{6}$ &
				\makecell{1.122e+3 \\ $\pm$8.943e+2} &
				\underline{\makecell{2.497e+2 \\ $\pm$1.609e+1}} &
				\makecell{3.919e+2 \\ $\pm$3.499e+1} &
				\makecell{5.214e+2 \\ $\pm$4.528e+1} &
				\makecell{5.729e+2 \\ $\pm$6.865e+1} &
				\makecell{1.006e+3 \\ $\pm$8.684e+2} &
				\makecell{3.379e+2 \\ $\pm$4.121e+1} &
				\makecell{3.428e+2 \\ $\pm$3.087e+1} &
				\textbf{\makecell{2.249e+2 \\ $\pm$1.733e+1}} \\
				$\mathcal{T}_{7}$ &
				\makecell{1.182e+3 \\ $\pm$1.015e+3} &
				\makecell{4.572e+2 \\ $\pm$5.786e+2} &
				\makecell{4.079e+2 \\ $\pm$2.047e+1} &
				\makecell{7.249e+2 \\ $\pm$5.632e+2} &
				\makecell{6.569e+2 \\ $\pm$3.398e+2} &
				\makecell{1.248e+3 \\ $\pm$1.033e+3} &
				\makecell{8.375e+2 \\ $\pm$8.195e+2} &
				\underline{\makecell{3.706e+2 \\ $\pm$3.569e+1}} &
				\textbf{\makecell{2.483e+2 \\ $\pm$8.558e+0}} \\
				$\mathcal{T}_{8}$ &
				\makecell{2.002e+3 \\ $\pm$1.214e+3} &
				\makecell{1.654e+3 \\ $\pm$1.049e+3} &
				\makecell{1.540e+3 \\ $\pm$9.585e+2} &
				\makecell{1.187e+3 \\ $\pm$9.914e+2} &
				\makecell{1.477e+3 \\ $\pm$9.215e+2} &
				\makecell{1.901e+3 \\ $\pm$1.179e+3} &
				\makecell{1.140e+3 \\ $\pm$1.052e+3} &
				\underline{\makecell{6.300e+2 \\ $\pm$6.961e+2}} &
				\textbf{\makecell{5.405e+2 \\ $\pm$7.052e+2}} \\
				$\mathcal{T}_{9}$ &
				\makecell{1.593e+3 \\ $\pm$1.131e+3} &
				\makecell{7.866e+2 \\ $\pm$7.858e+2} &
				\underline{\makecell{4.597e+2 \\ $\pm$5.666e+1}} &
				\makecell{6.039e+2 \\ $\pm$7.659e+1} &
				\makecell{1.325e+3 \\ $\pm$8.472e+2} &
				\makecell{1.705e+3 \\ $\pm$1.097e+3} &
				\makecell{8.367e+2 \\ $\pm$7.735e+2} &
				\makecell{7.917e+2 \\ $\pm$7.299e+2} &
				\textbf{\makecell{2.372e+2 \\ $\pm$1.417e+1}} \\
				$\mathcal{T}_{10}$ &
				\makecell{1.823e+3 \\ $\pm$1.044e+3} &
				\makecell{4.941e+2 \\ $\pm$5.591e+2} &
				\makecell{5.005e+2 \\ $\pm$3.361e+2} &
				\makecell{7.448e+2 \\ $\pm$3.612e+2} &
				\makecell{1.731e+3 \\ $\pm$8.654e+2} &
				\makecell{2.028e+3 \\ $\pm$8.949e+2} &
				\makecell{1.033e+3 \\ $\pm$8.609e+2} &
				\underline{\makecell{4.246e+2 \\ $\pm$3.384e+2}} &
				\textbf{\makecell{2.871e+2 \\ $\pm$3.337e+2}} \\
				$\mathcal{T}_{11}$ &
				\makecell{3.482e+3 \\ $\pm$8.906e+2} &
				\textbf{\makecell{1.230e+3 \\ $\pm$2.004e+2}} &
				\makecell{3.110e+3 \\ $\pm$6.810e+2} &
				\makecell{2.068e+3 \\ $\pm$6.013e+2} &
				\makecell{3.570e+3 \\ $\pm$4.091e+2} &
				\makecell{3.211e+3 \\ $\pm$8.862e+2} &
				\makecell{1.859e+3 \\ $\pm$9.007e+2} &
				\makecell{2.100e+3 \\ $\pm$8.135e+2} &
				\underline{\makecell{1.277e+3 \\ $\pm$8.204e+2}} \\
				$\mathcal{T}_{12}$ &
				\makecell{4.045e+3 \\ $\pm$3.519e+2} &
				\underline{\makecell{2.388e+3 \\ $\pm$9.204e+2}} &
				\makecell{3.845e+3 \\ $\pm$1.951e+2} &
				\makecell{2.619e+3 \\ $\pm$8.770e+2} &
				\makecell{4.056e+3 \\ $\pm$2.361e+2} &
				\makecell{4.183e+3 \\ $\pm$2.859e+2} &
				\makecell{3.026e+3 \\ $\pm$3.968e+2} &
				\makecell{3.228e+3 \\ $\pm$5.216e+2} &
				\textbf{\makecell{2.082e+3 \\ $\pm$9.189e+2}} \\
				$\mathcal{T}_{13}$ &
				\makecell{4.374e+3 \\ $\pm$1.262e+3} &
				\underline{\makecell{1.938e+3 \\ $\pm$9.041e+2}} &
				\makecell{3.989e+3 \\ $\pm$3.955e+2} &
				\makecell{3.007e+3 \\ $\pm$1.040e+3} &
				\makecell{4.621e+3 \\ $\pm$4.658e+2} &
				\makecell{4.824e+3 \\ $\pm$1.380e+3} &
				\makecell{2.882e+3 \\ $\pm$1.090e+3} &
				\makecell{3.498e+3 \\ $\pm$5.114e+2} &
				\textbf{\makecell{1.079e+3 \\ $\pm$8.001e+2}} \\
				$\mathcal{T}_{14}$ &
				\makecell{2.872e+3 \\ $\pm$1.189e+3} &
				\makecell{1.434e+3 \\ $\pm$9.130e+2} &
				\makecell{1.642e+3 \\ $\pm$5.567e+2} &
				\makecell{1.634e+3 \\ $\pm$4.497e+2} &
				\makecell{1.728e+3 \\ $\pm$4.054e+2} &
				\makecell{2.945e+3 \\ $\pm$1.053e+3} &
				\underline{\makecell{1.114e+3 \\ $\pm$5.911e+2}} &
				\makecell{1.188e+3 \\ $\pm$6.164e+1} &
				\textbf{\makecell{4.653e+2 \\ $\pm$3.986e+1}} \\
				$\mathcal{T}_{15}$ &
				\makecell{1.944e+3 \\ $\pm$6.455e+2} &
				\underline{\makecell{9.991e+2 \\ $\pm$1.359e+2}} &
				\makecell{1.701e+3 \\ $\pm$4.487e+2} &
				\makecell{1.650e+3 \\ $\pm$1.511e+2} &
				\makecell{1.723e+3 \\ $\pm$1.286e+2} &
				\makecell{2.607e+3 \\ $\pm$9.849e+2} &
				\makecell{1.136e+3 \\ $\pm$3.533e+2} &
				\makecell{1.286e+3 \\ $\pm$6.818e+1} &
				\textbf{\makecell{5.417e+2 \\ $\pm$4.102e+1}} \\
				$\mathcal{T}_{16}$ &
				\makecell{3.105e+3 \\ $\pm$1.025e+3} &
				\underline{\makecell{9.804e+2 \\ $\pm$1.429e+2}} &
				\makecell{3.092e+3 \\ $\pm$7.551e+2} &
				\makecell{2.221e+3 \\ $\pm$8.539e+2} &
				\makecell{3.619e+3 \\ $\pm$1.669e+2} &
				\makecell{3.276e+3 \\ $\pm$9.051e+2} &
				\makecell{1.657e+3 \\ $\pm$8.588e+2} &
				\makecell{1.597e+3 \\ $\pm$4.898e+2} &
				\textbf{\makecell{8.538e+2 \\ $\pm$5.407e+2}} \\
				
				\midrule
				
				$+$/$-$/$\approx$ 
				& 16/0/0 
				& 13/1/2
				& 16/0/0
				& 15/0/1
				& 16/0/0 
				& 16/0/0
				& 16/0/0
				& 15/0/1
				& N/A \\
				
				\bottomrule
			\end{tabular*}%
		}
	\end{table}

	\section{Experimental Results}
	\label{experiment}
	\subsection{Experimental Setup}
	\label{subsec:exp_setup}
	
\begin{figure}[t]
		\centering
		\includegraphics[width=0.85\linewidth]{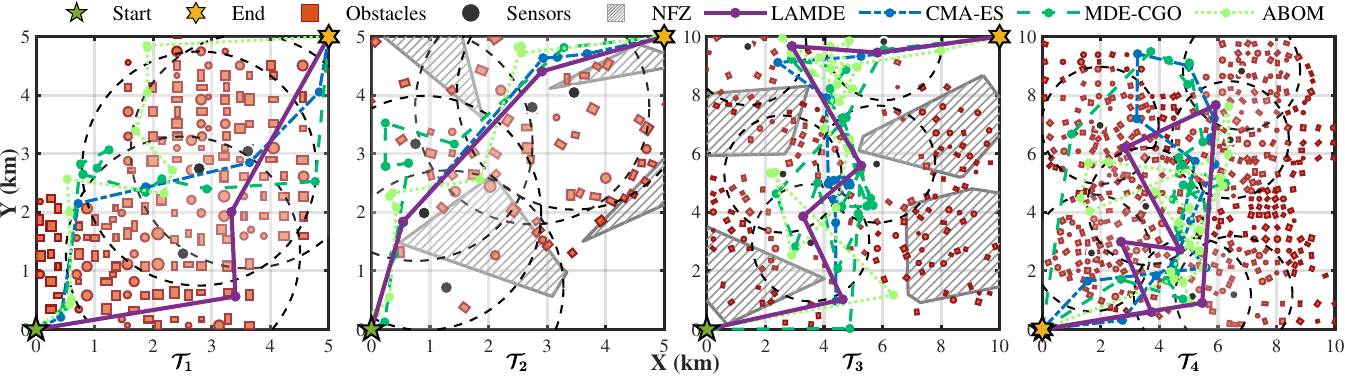}
		\caption{The best flying trajectories generated by LAMDE, CMA-ES, MDE-CGO, and ABOM on diverse maps $\mathcal{T}_1$, $\mathcal{T}_2$, $\mathcal{T}_{3}$, and $\mathcal{T}_{4}$.}
		\label{fig:exp1_P1}
	\end{figure}

	\noindent\textbf{Benchmark Suite.} The existing UAV path planning benchmarks~\cite{bai2025uav,sun20253d} lack the accurate and comprehensive modeling to capture the complex physical constraints of real-world UAV
	deployments. To bridge this gap, based on the basic modeling of 2D UAV path planning tasks, we construct a dedicated benchmark suite~(i.e., $\mathcal{D}_{test}$) comprising 16 WSN data collection settings for LAMDE and other compared baselines to do UAV path planning tasks. The generation of this suite follows our obstacle designs in Sec.~\ref{subsec:env_modeling}, spanning two map sizes---$5000
	\times 5000$ ($I_{\max}=10$) and $10000 \times 10000$
	($I_{\max}=20$), and with incremental obstacle density hence posing diverse difficulties for the underlying optimizers to solve. We provide 4 visualization examples of these task instances in Fig.~\ref{fig:exp1_P1}. An important point we must emphasize is that: in this paper, we strictly follow the challenging conditions in realistic scenario, that is, for each instance, the total number of evaluations allowed for solving it is limited to 10000.
	
	\noindent\textbf{Baselines.} We evaluate LAMDE against 8 representative baselines selected from both traditional evolutionary approaches and recent MetaBBO approaches. For the former, we adopt \textbf{CMA-ES}~\cite{cmaes}: a powerful global optimizer,
	\textbf{CMOCSO}~\cite{cmocso}: a competition-based and up-to-date optimizer for solving multi-objective multi-constraints problem, \textbf{L-STRDE}~\cite{lstrde}: the winner optimizer in IEEE CEC 2024 competition, and
	\textbf{MDE-CGO}~\cite{bai2025uav}: a matrix-based evolutionary optimizer tailored for WSN data collection scenarios. For the latter, we adopt \textbf{ABOM}~\cite{abom}, \textbf{LDE}~\cite{lde},
	\textbf{GLEET}~\cite{gleet}, and \textbf{RLDEAFL}~\cite{rldeafl}. These learning-assisted optimizers are representative ones that show competitive optimization performance in recent literature. For CMA-ES, L-STRDE and all MetaBBO baselines, since they are not originally designed for constrained optimization, we use classical penalty approach~\cite{penalty_handle}: $\hat{f}(\mathbf{x}) = f(\mathbf{x})+\delta \cdot \phi(\mathbf{x})$, to transform constraints into the objective function, where $\delta=10$. Despite these differences, all baselines follow their original settings in their papers.
	
	\noindent\textbf{Implementation Details.} LAMDE uses the pre-trained NeurELA feature extractor released by the original authors.\footnote{https://github.com/MetaEvo/Neur-ELA} Since LAMDE only requires a small number of real function evaluations to construct the proxy training set ($N_{probe}=100$), the remaining evaluation budget is reserved for test optimization. In contrast, other MetaBBO baselines without proxy training directly consume the full evaluation budget on $\mathcal{D}_{test}$. All experiments are independently repeated 25 times. We provide detailed implementation configurations in the supplementary materials, including the environment settings and training hyperparameters.
	
	\subsection{Systematic Performance Comparison}
	\label{subsec:exp_performance}
	We first compare the optimization performances of the baselines. For traditional optimizer baselines, we directly use them to optimize the given 16 UAV path planning instances. For learning-assisted MetaBBO baselines including our LAMDE, we test their trained versions on the instances. To compare baselines with a unified performance metric,  we use classical penalty approach~\cite{penalty_handle}~(smaller is better): $\hat{f}(\mathbf{x}) = f(\mathbf{x})+\delta \cdot \phi(\mathbf{x})$, and set $\delta$ as 10. The average optimization results and corresponding standard deviations across the 25 independent runs are reported in Table~\ref{tab:exp2}. The results in the table demonstrate that: 1) LAMDE significantly outperforms all compared baselines on almost all tested path planning instances. This underscores the effectiveness of each specific designs in our LAMDE as a whole system. More importantly, by meta learning on the training problem distribution, LAMDE could show robust generalization on unseen realistic problems. 2) LAMDE only underperforms the powerful CMA-ES optimizer on two cases $\mathcal{T}_4$ and $\mathcal{T}_{11}$ with slight performance gap. This is mainly due to the performance computation focuses on the statistical average across multiple runs. As a powerful optimizer, it is not a surprise at all the CMA-ES could perform robustly on some realistic problems. However, as we presented in Fig.~\ref{fig:exp1_P1}, the best trajectories our LAMDE finds are superior to CMA-ES and other baselines. 3) LAMDE achieves state-of-the-art performance within the MetaBBO baselines. A major reason behind is the training data preparation we propose in Sec.~\ref{sec:landscape-aware}. Given this scheme, we can augment the training data without loss of distribution alignment. In contrast, the compared MetaBBO baselines are limited by the target optimization tasks and restricted evaluation resources. This indicates that LAMDE is more capable of solving optimization challenges in the wild. Note that this is a step forward for existing MetaBBO paradigms. 4) An up-to-date MetaBBO baseline ABOM shows competitive results on such realistic tasks, while it still performs worse than our LAMDE. What we want to discuss here is the design philosophy of ABOM: training online policy along the actual optimization process with adaptive self-supervised learning. Compared to the other three MetaBBO baselines, ABOM performs better, which indicates that such online learning is also a promising way to address realistic challenging optimization tasks. 5) Generally, we validate the learning potential for automating algorithm design in optimization tasks. While the ranks of the four traditional baselines show clear variance, our LAMDE ranks the first place consistently with automated algorithm configuration. 6) A specific case we would like to mention here is the comparison between CMOCSO~(a tailored constraint optimizer) and CMA-ES~(an unconstrained optimizer with penalty objective). CMA-ES beats CMOCSO in every tested instance, with significant performance gap. This more or less reflects the existing status in human-crafted evolutionary optimizers. Once they are used in unseen scenarios, it is extremely hard to ensure the claimed optimization robustness. This further underscores the importance of exploring more on learning-assisted optimization techniques such as MetaBBO.

	We further visualize the best flying trajectories found by the top-4 optimizers across 25 independent runs in Fig.~\ref{fig:exp1_P1} to illustrate the actual effectiveness of LAMDE. We could observe that the three baselines frequently produce irregular paths with redundant detours, whereas LAMDE generates more concise and smoother routes, owing to its generalized meta-level policy for flexible search and the variable-length encoding in the low-level optimization to ensure ability to adaptively prune unnecessary hover points. We provide the visualization results of the rest 12 cases at our project. In addition, to further investigate the searching behavior of LAMDE, we decompose the optimization metric into objective value and constraint violation value, and analyze their search behaviors provided in the supplementary materials.
	
	\begin{figure}[t]
		\centering
		\includegraphics[width=0.75\linewidth]{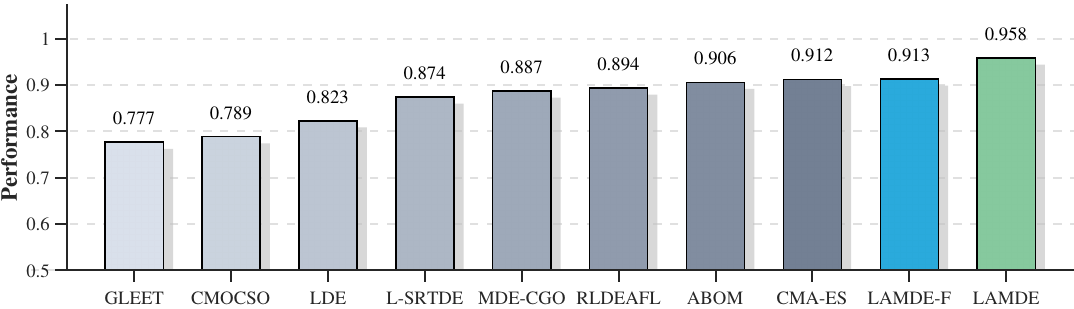}
		\caption{Comparison of performance between LAMDE-F and baselines}
		\label{fig:exp2_P1}
	\end{figure}
	
	\subsection{Ablation Studies}
	\label{subsec:ablation_retrieval}
	We would like to again emphasize two major contributions in LAMDE's methodology: i) the \textbf{variable-length encoding strategy} used to facilitate searching flexibility in the low-level optimization. ii) the \textbf{landscape-aware training data augmentation scheme} to enlarge the training scale of the meta-level policy. In this ablation section, we aim to demonstrate the effectiveness of these two core contributions through rigorous experiments.
	
	First, we focus on the variable-length encoding strategy. Specifically, we ablate this strategy in LAMDE by replacing it with a fixed gene encoding, termed as LAMDE-F. This ablated variant enforces the low-level DE optimizer to search feasible solutions with $I_{\max}$ hover points. We plot the average normalized performances\footnote{The normalization is performed in a per-run manner. In each run, the final performance is normalized by the initial performance. Then performances of 25 runs over all 16 test cases are aggregated.} of all previous baselines and LAMDE-F in Fig~\ref{fig:exp2_P1}. The results indicate that: 1)  LAMDE achieves consistently higher and more stable returns than LAMDE-F across all 16 instances, confirming the benefit of adaptive trajectory pruning. 2) LAMDE-F still slightly outperforms all selected baselines, this on the other hand validates the learning of LAMDE is effective.

	\begin{table}[t]
		\centering
		\renewcommand{\arraystretch}{0.8}
		\caption{Ablation of retrieval strategies on 16 UAV instances.}
		\label{tab:ablation}
		\resizebox{\columnwidth}{!}{
			\begin{tabular}{lccccc}
				\toprule
				ID & LAMDE-R & LAMDE-E & LAMDE-L & LAMDE-O & \textbf{LAMDE} \\
				\midrule
				$\mathcal{T}_{1}$ &
				1.58e+3 $\pm$ 8.02e+2 &
				1.62e+3 $\pm$ 6.13e+2 &
				1.69e+3 $\pm$ 5.52e+2 &
				1.36e+3 $\pm$ 7.74e+2 &
				\textbf{1.36e+3 $\pm$ 7.79e+2} \\
				$\mathcal{T}_{2}$ &
				1.82e+2 $\pm$ 7.78e+0 &
				1.82e+2 $\pm$ 8.95e+0 &
				1.82e+2 $\pm$ 9.02e+0 &
				1.84e+2 $\pm$ 1.62e+1 &
				\textbf{1.82e+2 $\pm$ 7.57e+0} \\
				$\mathcal{T}_{3}$ &
				6.93e+2 $\pm$ 4.49e+2 &
				8.71e+2 $\pm$ 6.19e+2 &
				5.52e+2 $\pm$ 4.74e+1 &
				6.77e+2 $\pm$ 3.15e+2 &
				\textbf{5.51e+2 $\pm$ 6.69e+1} \\
				$\mathcal{T}_{4}$ &
				2.48e+3 $\pm$ 3.95e+2 &
				2.50e+3 $\pm$ 5.91e+2 &
				2.62e+3 $\pm$ 1.80e+2 &
				2.45e+3 $\pm$ 5.49e+2 &
				\textbf{2.39e+3 $\pm$ 6.12e+2} \\
				$\mathcal{T}_{5}$ &
				1.96e+2 $\pm$ 6.62e+0 &
				1.94e+2 $\pm$ 8.91e+0 &
				1.94e+2 $\pm$ 4.32e+0 &
				1.92e+2 $\pm$ 5.73e+0 &
				\textbf{1.93e+2 $\pm$ 5.95e+0} \\
				$\mathcal{T}_{6}$ &
				2.88e+2 $\pm$ 5.39e+1 &
				2.78e+2 $\pm$ 7.92e+1 &
				2.28e+2 $\pm$ 1.29e+1 &
				2.22e+2 $\pm$ 8.93e+0 &
				\textbf{2.25e+2 $\pm$ 1.73e+1} \\
				$\mathcal{T}_{7}$ &
				2.77e+2 $\pm$ 4.41e+0 &
				2.48e+2 $\pm$ 1.20e+1 &
				3.10e+2 $\pm$ 3.15e+2 &
				2.50e+2 $\pm$ 1.08e+1 &
				\textbf{2.48e+2 $\pm$ 8.56e+0} \\
				$\mathcal{T}_{8}$ &
				7.23e+2 $\pm$ 8.36e+2 &
				6.45e+2 $\pm$ 7.70e+2 &
				6.74e+2 $\pm$ 7.55e+2 &
				7.13e+2 $\pm$ 8.01e+2 &
				\textbf{5.41e+2 $\pm$ 7.05e+2} \\
				$\mathcal{T}_{9}$ &
				3.06e+2 $\pm$ 3.34e+2 &
				2.51e+2 $\pm$ 2.96e+1 &
				2.52e+2 $\pm$ 2.23e+1 &
				2.39e+2 $\pm$ 1.80e+1 &
				\textbf{2.37e+2 $\pm$ 1.42e+1} \\
				$\mathcal{T}_{10}$ &
				3.32e+2 $\pm$ 3.76e+2 &
				2.97e+2 $\pm$ 3.92e+2 &
				2.90e+2 $\pm$ 3.23e+2 &
				3.54e+2 $\pm$ 4.69e+2 &
				\textbf{2.87e+2 $\pm$ 3.34e+2} \\
				$\mathcal{T}_{11}$ &
				1.42e+3 $\pm$ 7.31e+2 &
				2.38e+3 $\pm$ 7.11e+2 &
				1.53e+3 $\pm$ 9.48e+2 &
				1.36e+3 $\pm$ 8.92e+2 &
				\textbf{1.28e+3 $\pm$ 8.20e+2} \\
				$\mathcal{T}_{12}$ &
				2.56e+3 $\pm$ 3.90e+2 &
				2.55e+3 $\pm$ 8.22e+2 &
				2.35e+3 $\pm$ 7.19e+2 &
				2.36e+3 $\pm$ 6.91e+2 &
				\textbf{2.08e+3 $\pm$ 9.19e+2} \\
				$\mathcal{T}_{13}$ &
				1.14e+3 $\pm$ 9.11e+2 &
				1.56e+3 $\pm$ 9.00e+2 &
				1.23e+3 $\pm$ 9.13e+2 &
				1.51e+3 $\pm$ 8.76e+2 &
				\textbf{1.08e+3 $\pm$ 8.00e+2} \\
				$\mathcal{T}_{14}$ &
				4.89e+2 $\pm$ 4.42e+1 &
				4.70e+2 $\pm$ 4.13e+1 &
				4.84e+2 $\pm$ 4.19e+1 &
				4.70e+2 $\pm$ 5.84e+1 &
				\textbf{4.65e+2 $\pm$ 3.99e+1} \\
				$\mathcal{T}_{15}$ &
				5.65e+2 $\pm$ 7.44e+1 &
				5.81e+2 $\pm$ 7.28e+1 &
				6.27e+2 $\pm$ 3.08e+2 &
				6.69e+2 $\pm$ 2.96e+2 &
				\textbf{5.42e+2 $\pm$ 4.10e+1} \\
				$\mathcal{T}_{16}$ &
				8.66e+2 $\pm$ 5.72e+2 &
				8.90e+2 $\pm$ 4.73e+2 &
				8.93e+2 $\pm$ 5.81e+2 &
				1.24e+3 $\pm$ 7.99e+2 &
				\textbf{8.54e+2 $\pm$ 5.41e+2} \\
				\bottomrule
			\end{tabular}
		}
	\end{table}

	Second, we validate the importance of the landscape-aware module by preparing four ablated baselines: (i)~\textbf{LAMDE-R}, which samples synthetic training tasks at random without any landscape guidance; (ii)~\textbf{LAMDE-E}, which replaces NeurELA with classical ELA features~\cite{ela} while retaining the same matching procedure; and (iii)~\textbf{LAMDE-L}, which performs the landscape locally: instead of locating synthetic functions with distributional similarity, we locate a group of similar synthetic instances for each target task; (iv)~\textbf{LAMDE-O}: we entirely drop the prepared training set and train LAMDE as the original setting in existing MetaBBO baselines. Table~\ref{tab:ablation} reports the results across all 16 instances. LAMDE-R performs worst overall, confirming that unguided sampling produces training distributions misaligned with the target domain. LAMDE-E improves upon random sampling, yet remains clearly inferior to our LAMDE, exposing the limited representational capacity of handcrafted features when characterizing complex constrained landscapes. LAMDE-L narrows the gap further, but its instance-wise retrieval is inconsistent with the learning philosophy of MetaBBO: learning over problem distribution. LAMDE-O also performs significantly worse than LAMDE, which further validates the bottlenecks of existing MetaBBO paradigms in solving realistic problems. Our LAMDE, by aligning synthetic tasks through global centroid matching, captures coherent domain-level structure and achieves the best performance, validating that both the learned feature representation and the holistic matching strategy are indispensable for effective zero-shot generalization.

	\subsection{In-depth Analysis}
	\label{subsec:impact_encoding}

	A key observation in Fig.~\ref{fig:exp2_P1} is that, with the enhancement of the variable-length encoding strategy in the low-level optimization, LAMDE performs significantly better learning effectiveness and generalized optimization performance. We think this is an important experimental result and it deserves more exploration. To this end, in this section we provide a series of in-depth analysis on this aspect. Recall that
	to isolate the contribution of the proposed variable-length encoding, we construct a variant, \textbf{LAMDE-F}, that retains the landscape aware meta-policy but removes the Variable-Length Encoding Strategy. The policy is trained by the same setting as LAMDE. We use this variant and our LAMDE for subsequent analysis.
	
\begin{figure}[t]
		\centering
		\includegraphics[width=0.6\linewidth]{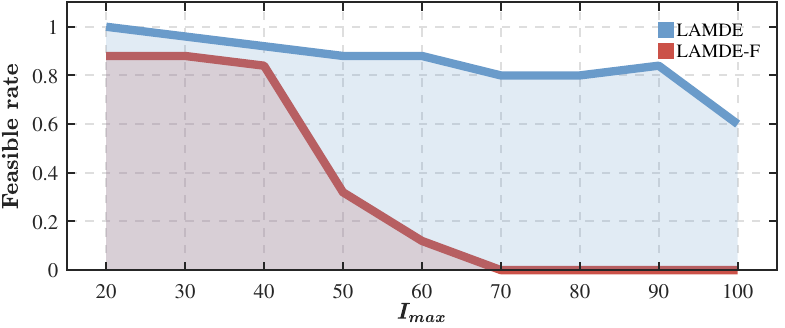}
		\caption{Average feasible rates under increasing $I_{\max}$ across test set.}
		\label{fig:exp2_P3}
	\end{figure}

	Fig.~\ref{fig:exp2_P3} illustrates the feasible rates of LAMDE and LAMDE-F as the maximum problem dimension~(i.e., $I_{max}$ hover points) increases. These rates are computed by counting the proportion of feasible solutions in the last-generation population across all 16 target tasks and 25 independent runs. We could observe that as the $I_{max}$ increases from 20 to 100, LAMDE-F deteriorates rapidly and fails to produce feasible solutions once $I_{\max} \ge 70$, burdened by optimizing an expanding set of redundant hover points. In contrast, with our proposed variable-length encoding strategy, LAMDE maintains a high feasibility rate as the decision variable dimension increases. Specifically, it ensures a feasible rate above 0.6 even at $I_{\max}=100$, demonstrating strong robustness and is preferable in practical complex optimization environments. We believe this is achieved by the added decision dimension~(i.e., the activation value) in the solution. This added dimension provides the low-level optimizer a more fine-grained interface to determine the sparsity of the solution. Furthermore, such additional dynamics of the low-level optimization could provide a more informative environmental dynamics for the meta-level policy to learn more preferable algorithm configuration decisions, which in turn helps enhance the overall optimization performance. Due to space constraints, further analyses regarding the algorithmic behaviors and the broader applicability of this variable-length strategy across other optimizers are detailed in the supplementary materials.
	
	\section{Conclusion}
	\label{sec:conclusion}
	In this paper, we propose a systematic MetaBBO framework LAMDE to address realistic UAV path planning~(with challenging obstacle layout) in WSN data collection tasks. In particular, we first re-formulate the optimization model in such planning scenarios with city-level obstacles and various running constraints. Then we dive into the design space of the MetaBBO paradigm. At the meta level, we propose an landscape-aware automatic training data preparation scheme to mitigate the potential distribution shift issue in existing MetaBBO training paradigms. At the lower level, we equip EC optimizers with a variable-length encoding strategy to facilitate solving flexibility of low-level optimization process. Experiments on diverse UAV missions show that LAMDE consistently outperforms both classical and learning-based baselines. Fuether in-depth analysis underscore the effectiveness of our designs. We also mark potential future works that could improve LAMDE further: i) 3D UAV environmental setting. ii) Online dynamic planning in noisy environment. iii) Multi-UAV co-evolution. We believe LAMDE is not only an effective solver for challenging UAV path planning tasks, but also a practical guideline for those who develop MetaBBO for realistic problems.
	
    \bibliographystyle{IEEEtran}
	\bibliography{reference}
	
\end{document}